%% file: root.tex
\documentclass[letterpaper, 10 pt, conference]{ieeeconf}  

\usepackage{cite}
\usepackage{amsmath,amssymb,amsfonts}
\usepackage{algorithmic}
\usepackage{graphicx}
\usepackage{subcaption}
\usepackage{textcomp}
\usepackage{xcolor}
\usepackage{float} 
\usepackage{placeins}
\usepackage{multirow, makecell, graphicx, amssymb} 
\usepackage{lipsum} 
\usepackage{caption}
\usepackage{booktabs} 

\IEEEoverridecommandlockouts                              

\title{\LARGE \bf
PanoDP: Learning Collision-Free Navigation \\with Panoramic Depth and Differentiable Physics
}

\author{Hao Zhong$^{1}$, Pei Chi$^{2}$, Jiang Zhao$^{2}$, Shenghai Yuan$^{1*}$, Xuyang Gao$^{2}$, Thien-Minh Nguyen$^{3}$, and Lihua Xie$^{1}$%
\thanks{$^{1}$Hao Zhong, Shenghai Yuan, and Lihua Xie are with the School of Electrical and Electronic Engineering, Nanyang Technological University, Singapore.}%
\thanks{$^{2}$Pei Chi, Jiang Zhao, and Xuyang Gao are with Beihang University, Beijing, China.}%
\thanks{$^{3}$Thien-Minh Nguyen is with The University of Queensland, Australia.}%
\thanks{$^{*}$Corresponding author: Shenghai Yuan.}%
}

\begin{document}

\maketitle
\thispagestyle{empty}
\pagestyle{empty}

\begin{abstract}
Autonomous collision-free navigation in cluttered environments requires safe decision-making under partial observability with both static structure and dynamic obstacles. We present \textbf{PanoDP}, a communication-free learning framework that combines four-view panoramic depth perception with differentiable-physics-based training signals. PanoDP encodes panoramic depth using a lightweight CNN and optimizes policies with dense differentiable collision and motion-feasibility terms, improving training stability beyond sparse terminal collisions. We evaluate PanoDP on a controlled ring-to-center benchmark with systematic sweeps over agent count, obstacle density/layout, and dynamic behaviors, and further test out-of-distribution generalization in an external simulator (e.g., AirSim). Across settings, PanoDP increases collision-free and completion rates over single-view and non-physics-guided baselines under matched training budgets, and ablations (view masking, rotation augmentation) confirm the policy leverages 360-degree information. Code will be open source upon acceptance. 
\end{abstract}

\section{Introduction}
\label{sec:intro}

Autonomous collision-free navigation has become a highly active area because it is a prerequisite for deploying robots and aerial vehicles in crowded real-world environments with both static structures and dynamic obstacles, including other robots. In multi-robot settings, a practical requirement often arises: the system should remain functional without explicit inter-robot communication, since communication can be unavailable, unreliable, delayed, bandwidth-limited, or operationally undesirable. The core issue is that interaction dynamics are inherently coupled, yet each robot must make safe decisions from local sensing only, which can easily lead to collisions or deadlocks when observability is limited.

\textbf{Existing} work on decentralized, communication-free collision avoidance spans classical reciprocal interaction rules, optimization-based safety layers, and learning-based policies.
Reciprocal collision avoidance, such as ORCA, provides principled communication-free interaction rules for multi-agent navigation~\cite{van2010optimal}, but it can become brittle or overly conservative in dense scenes and under perception imperfections.
Control-barrier-certificate style methods provide formal safety guarantees by constraining a nominal controller~\cite{borrmann2015control}, yet they often assume reliable access to state or accurate estimates and can be restrictive when perception is uncertain or when many agents interact tightly.
Learning-based approaches, including decentralized sensor-level deep RL, can map onboard observations to actions and demonstrate scalability~\cite{long2018towards}, but they are often sensitive to reward design and hyperparameters, and their generalization across obstacle layouts, densities, and interaction patterns can be inconsistent.


The \textbf{challenges} are threefold. First, decentralized navigation is partially observable: limited field-of-view sensors create blind spots and occlusions, so hazards can approach from the side or rear and other robots can be intermittently hidden, leading to collisions and deadlocks. Second, collision avoidance provides sparse supervision, which makes training slow and brittle; stable learning requires dense signals that reflect safety and feasibility throughout the trajectory rather than only at terminal events. Third, expanding sensing coverage via multi-view or panoramic depth can reduce blind spots but introduces practical compute and memory costs, and improved observability alone does not resolve the optimization difficulty of learning safe behaviors under mixed static-dynamic interactions.

To address these challenges, we propose \textbf{PanoDP}, a communication-free navigation framework that couples panoramic depth perception with differentiable-physics-based training signals.
Panoramic depth provides 360-degree coverage and acts as a substitute for neighbor-state exchange, allowing each robot to anticipate hazards beyond a front sensor using onboard sensing.
Meanwhile, a differentiable physics module provides shaped safety and feasibility objectives via differentiable collision and motion-feasibility terms~\cite{qiao2020scalable}, enabling end-to-end optimization by backpropagating signals rather than relying solely on sparse terminal collisions.
We do not claim a new differentiable physics formulation; instead, we show that integrating such differentiable-physics signals with panoramic depth yields denser, more stable training and improves decentralized navigation performance without explicit inter-robot communication. Our Contribution can be summarized as follows:
\begin{itemize}
    \item We formulate decentralized, communication-free multi-robot collision-free navigation under partial observability, where each agent acts solely from local panoramic depth observations.
    \item We propose \textbf{PanoDP}, coupling a lightweight panoramic depth encoder with differentiable-physics-based objectives for collision avoidance and motion feasibility, providing dense and stable training signals.
    \item We introduce a controlled ring-to-center benchmark with systematic sweeps over agent count, obstacle density/layout, and dynamic behaviors, and we validate performance via matched-budget comparisons and targeted ablations.
\end{itemize}

\section{Related Work}
\label{sec:related_work}

\noindent\textbf{Multi-Robot Navigation.}
Prior work studies collision avoidance under centralized coordination and decentralized execution. In the communication-free setting, each agent acts from local observations under interactions. Classical reciprocal and velocity-obstacle-style rules~\cite{van2010optimal, yasuda2023safe, wang2024apf} are efficient and interpretable, but they can become conservative or brittle in dense clutter, with complex interaction patterns, or under sensing uncertainty. Optimization-based safety~\cite{borrmann2015control, fernandez2023distributed, mestres2024distributed, gonccalves2024safe,park2024resilient} can enforce constraints on a nominal controller, but they rely on accurate state or estimates and may be restrictive when many agents interact. Decentralized learning-based policies~\cite{long2018towards} can capture interaction patterns and implicit coordination, yet training can be sensitive and generalization across layouts and agent densities remains inconsistent.

\noindent\textbf{Panoramic and Multi-View Perception.}
Multi-view and panoramic sensing~\cite{liu2024omninxt, gattaux2025route, ge2025airsim360, shah2017airsim} have been explored to reduce blind spots and occlusions in navigation. For multi-robot systems, wider angular coverage can partially substitute for communication by enabling observation of agents in directions, reducing reliance on neighbor-state exchange. At the same time, panoramic encoding increases compute and observation dimensionality, which can slow training and complicate deployment. It is also non-trivial to determine whether learned policies exploit 360-degree information rather than defaulting to a front-view prior, especially when training data is biased. View masking, partial-view reductions, and rotation-based tests are therefore important to validate genuine panoramic utilization.

\noindent\textbf{Differentiable Physics for Safety and Feasibility.}
Differentiable physics provides gradients through dynamics and contacts~\cite{qiao2020scalable}, enabling dense objectives related to collision avoidance and motion feasibility. Such structured signals can stabilize optimization compared with sparse terminal collision supervision and can better align training with physically feasible behaviors. Differentiable simulation has been applied to trajectory optimization, system identification, and policy learning~\cite{zhang2025learning, zhang2025diffaero}, often improving sample efficiency and reducing the gap between training objectives and execution constraints. However, these methods are typically used either for single-agent control~\cite{loquercio2021learning, zhang2025threat, huang2025towards, lee2025quadrotor} or as a generic training tool. In our setting, differentiable physics is used specifically to supply safety and feasibility terms for communication-free decentralized navigation, rather than to introduce a new physics model.

\noindent\textbf{Positioning.}
PanoDP targets communication-free decentralized navigation under partial observability by combining panoramic depth perception with differentiable-physics-based training objectives. Panoramic depth reduces blind spots without message passing, while physics-informed objectives provide dense safety and feasibility supervision that mitigates the brittleness of learning from sparse collisions. This combination yields a scalable framework for safe multi-robot navigation in mixed static and dynamic environments.


\section{Method}
\label{sec:method}

\subsection{ Formulation}
\label{sec:problem}
We address decentralized collision-free navigation for $N$ quadrotors operating in cluttered, dynamic environments. Each agent $i$ evolves on $\mathrm{SE}(3)$ with state $\mathbf{q}^{i}_{t}=\bigl(\mathbf{p}^{i}_{t},\,R^{i}_{t},\,\mathbf{v}^{i}_{t}\bigr)\in\mathbb{R}^{3}\!\times\!\mathrm{SO}(3)\!\times\!\mathbb{R}^{3}$, where $\mathbf{p}^{i}_{t}$ is the position, $R^{i}_{t}$ the body orientation, and $\mathbf{v}^{i}_{t}$ the linear velocity.
The agent has \emph{no access to global map, neighbor identities,
or privileged planner state}.
All it receives at each time step is (i)~onboard depth images from its
cameras and (ii)~a compact body-frame feature vector.
We seek a shared decentralized policy $\pi_\theta$ that maps each
agent's local observation to a continuous control command
$\mathbf{u}^{i}_{t}\!\in\!\mathbb{R}^{6}$, executed by the quadrotor
actuation model.
The objective is to maximize task completion (reaching assigned goals)
while minimizing collisions and ensuring smooth trajectories.

\noindent\emph{Remark} (Scalability).
Because each agent executes $\pi_\theta$ using only its own onboard
depth and IMU data, no inter-agent communication channel is required.
The per-agent computational cost is $O(1)$—independent of the swarm
size $N$—so adding a new quadrotor amounts to loading the same
policy weights without retraining or reconfiguration.
We validate this property empirically by scaling the same policy
from $N\!=\!64$ to $N\!=\!512$ without performance degradation.

\subsection{Panoramic Depth Observation}
\label{sec:panoramic_depth}
A forward-only depth camera covers at most $\sim\!90^{\circ}$ of the
agent's surroundings, creating blind spots during lateral or
backward maneuvers.
PanoDP replaces this narrow field of view with a full
$360^{\circ}$ panoramic depth representation.

\noindent\textbf{Multi-camera input.}\;
At each step $t$, four depth images are captured at
cardinal yaw offsets
$\phi_{k}\!\in\!\Phi=\{0,\tfrac{\pi}{2},\pi,\tfrac{3\pi}{2}\}$
from the body frame, each with horizontal field of view
$\alpha$:
\begin{equation}
\bigl\{\mathbf{D}^{i,\phi_{k}}_{t}\bigr\}_{k=1}^{4},
\qquad
\mathbf{D}^{i,\phi_{k}}_{t}\in\mathbb{R}^{H_{0}\times W_{0}}.
\end{equation}
Because $\alpha>90^{\circ}$, adjacent cameras overlap, providing
redundancy at the seam boundaries.

\noindent\textbf{Equirectangular projection.}\;
A stitching operator $\mathcal{S}$ composites the four views into
a single equirectangular panorama
$\mathbf{P}^{i}_{t}\!\in\!\mathbb{R}^{H\times W}$.
Every pixel $(u,v)$ of $\mathbf{P}$ corresponds to a spherical
direction $(\theta,\varphi)$:
\begin{equation}
\theta = \frac{2\pi\, u}{W},\qquad
\varphi = \left(\frac{v}{H}-\frac{1}{2}\right)\alpha_{v},
\label{eq:equirect_coord}
\end{equation}
where $\theta\!\in\![0,2\pi)$ is azimuth and $\varphi$ is elevation
within the field of view $\alpha_{v}$.
For each camera $k$ with yaw center $\phi_{k}$ and focal length
$f\!=\!W_{0}/(2\tan\frac{\alpha}{2})$, the spherical direction
$(\theta,\varphi)$ is expressed in camera coordinates and
reprojected to a location $(c_x, c_y)$ in $\mathbf{D}^{i,\phi_k}_t$
via the pinhole model:
\begin{align}
c_x &= f\,\tan(\theta-\phi_k)
      + \tfrac{W_0}{2},
\label{eq:reproj_cx}\\[2pt]
c_y &= f\,\frac{\sin\varphi}{\cos\varphi\,\cos(\theta\!-\!\phi_k)}
      + \tfrac{H_0}{2}.
\label{eq:reprojection}
\end{align}
Bilinear interpolation retrieves sub-pixel depth values.
In overlap regions, the results from adjacent cameras are blended
using a cosine-squared azimuthal weight
\begin{equation}
w_{k}(\theta)=\cos^{2}\!\!\left(\frac{\pi\,(\theta-\phi_{k})}{\alpha}\right),
\label{eq:blend_weight}
\end{equation}
which smoothly fades each camera's contribution near its field-of-view
boundary.
The final panorama pixel is
$\mathbf{P}[u,v]=\sum_{k}w_{k}\,\mathbf{D}^{(k)}[c_y,c_x]\,\big/\,\sum_{k}w_{k}$.
\begin{figure}[t]
\vspace{8pt}
\centering
\includegraphics[width=\columnwidth]{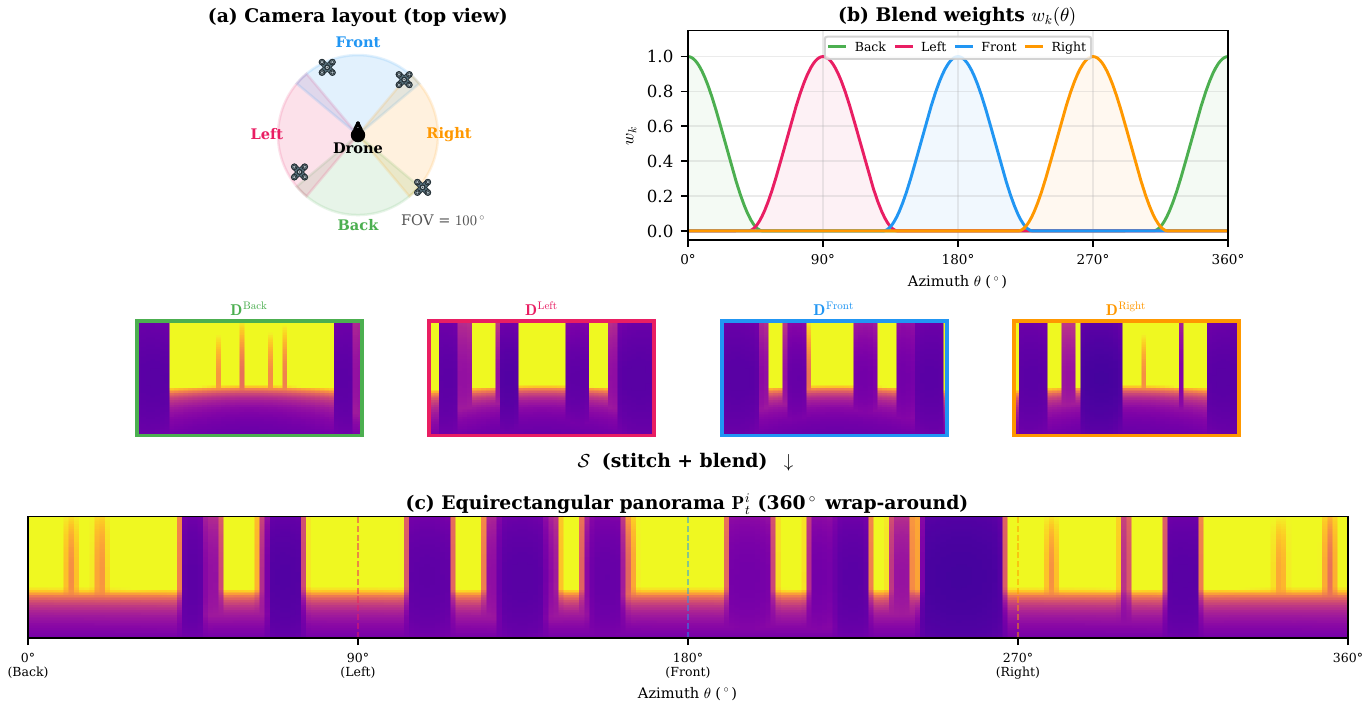}
\vspace{-18pt}
\caption{Panoramic depth construction.
(a)~Four cameras with $100^{\circ}$ FOV cover the full azimuth with
overlapping regions.
(b)~Cosine-squared blend weights $w_k(\theta)$ ensure smooth
transitions.
(c)~The resulting equirectangular panorama $\mathbf{P}^i_t$ wraps
continuously at $0^{\circ}/360^{\circ}$.}
\label{fig:panorama_stitching}
\vspace{-15pt}
\end{figure}
The horizontal axis of $\mathbf{P}$ wraps continuously around the full
azimuth: column $u$ encodes a unique bearing, so the panorama is a
compact representation of obstacle proximity in every direction.

\noindent\textbf{Depth preprocessing.}\;
A normalization pipeline transforms $\mathbf{P}$ into the network
input:
clamping to a bounded range $[d_{\min},d_{\max}]$, an inverse-depth
transform $\psi(d)=\beta/d - \gamma$, additive Gaussian noise for
sim-to-real regularization, and max-pool downsampling:
\begin{equation}
\tilde{\mathbf{P}}^{i}_{t}
= \mathrm{MaxPool}\!\bigl(\,\psi\bigl(\mathrm{clamp}(\mathbf{P}^{i}_{t})\bigr)
  + \epsilon\bigr),
\;\; \epsilon\sim\mathcal{N}(0,\sigma^{2}).
\label{eq:preprocess}
\end{equation}

\noindent\textbf{Body-frame feature vector.}\;
Alongside the panorama, each agent receives a 10-dimensional
onboard state
$\mathbf{s}^{i}_{t}\!\in\!\mathbb{R}^{10}$:
\begin{equation}
\mathbf{s}^{i}_{t}
=\bigl[\,\underbrace{\mathbf{v}^{i}_{\mathrm{local}}}_{\mathbb{R}^{3}}\;;\;
       \underbrace{\mathbf{v}^{i\to g}_{\mathrm{cmd}}}_{\mathbb{R}^{3}}\;;\;
       \underbrace{R^{i}_{t}\mathbf{e}_{z}}_{\mathbb{R}^{3}}\;;\;
       \underbrace{m^{i}_{t}}_{\mathbb{R}^{1}}\,\bigr],
\label{eq:state_vec}
\end{equation}
where $\mathbf{v}^{i}_{\mathrm{local}}$ is the current velocity
in a yaw-aligned body frame,
$\mathbf{v}^{i\to g}_{\mathrm{cmd}}$ is the speed-clamped
goal-directed velocity command in the same frame,
$R^{i}_{t}\mathbf{e}_{z}$ encodes the current pitch and roll
attitude, and $m^{i}_{t}$ is a scalar safety margin.
The yaw-aligned frame $R_{\mathrm{yaw}}$ is constructed by
projecting the body forward axis onto the horizontal plane,
thereby decoupling heading from tilt.
All components are derived from the agent's own onboard
state estimate, requiring no inter-agent communication.

\subsection{Circular Panoramic Encoder}
\label{sec:circular_conv}
Panoramic images are periodic along the horizontal (azimuth) axis:
the leftmost and rightmost columns represent adjacent bearings.
Standard zero-padded convolutions introduce artificial boundary
discontinuities at the panorama edges, corrupting features that
straddle the $0^{\circ}/360^{\circ}$ seam.
PanoDP replaces horizontal zero-padding with \emph{circular padding}:
the input is wrapped around so that the first $k$ columns reappear
after the last column (and vice versa), where $k$ is the kernel
half-width.
Formally, let $\mathcal{C}_{k}$ denote circular padding width $k$,
\begin{equation}
\mathcal{C}_{k}(\tilde{\mathbf{P}})[h,w]
= \tilde{\mathbf{P}}\bigl[h,\;(w \bmod W)\bigr],
\end{equation}
and vertical padding remains zero-valued.
Every horizontal convolution operates on
$\mathcal{C}_{k}(\tilde{\mathbf{P}})$, yielding a seamless
360\textdegree{} receptive field.
The visual embedding is
\begin{equation}
\mathbf{z}^{i}_{t}
= E_{\theta}\!\bigl(\tilde{\mathbf{P}}^{i}_{t}\bigr)
\;\in\;\mathbb{R}^{d_{z}},
\label{eq:encoder}
\end{equation}
where $E_{\theta}$ is a lightweight CNN whose every convolutional
layer uses circular horizontal padding, followed by a linear
projection to $d_z$ dimensions.
Architecture details are given in Sec.~\ref{sec:experiments}.
With circular padding, the encoder's receptive field uniformly covers
all bearings regardless of the agent's yaw; no azimuth direction
receives preferential treatment.

\subsection{Recurrent Policy with Temporal Memory}
\label{sec:gru_policy}
A single panoramic frame encodes the distance field but
contains no information about obstacle motion.
In dynamic swarms where neighboring drones move at comparable
speeds, temporal context is necessary to infer relative velocities
from consecutive depth observations.
We therefore augment the policy with a GRU-based recurrent module:
\begin{align}
\mathbf{e}^{i}_{t} &= W_{s}\,\mathbf{s}^{i}_{t},
\label{eq:state_embed}\\
\mathbf{h}^{i}_{t} &= \mathrm{GRU}\!\bigl(\mathbf{h}^{i}_{t-1},\;
                       \sigma(\mathbf{z}^{i}_{t}+\mathbf{e}^{i}_{t})\bigr),
\label{eq:gru}\\
\mathbf{u}^{i}_{t} &= W_{u}\,\sigma(\mathbf{h}^{i}_{t}).
\label{eq:action}
\end{align}
\indent The visual feature $\mathbf{z}^{i}_{t}$ and the state embedding
$\mathbf{e}^{i}_{t}$ are summed, passed through an activation
$\sigma$, and gated by the GRU.
The hidden state $\mathbf{h}^{i}_{t}$ accumulates temporal
information across frames, enabling the agent to implicitly
estimate its velocity and leverage motion context from
consecutive depth observations for smoother control.
An auxiliary head predicts the agent's velocity as a
self-supervised signal (Sec.~\ref{sec:diff_physics}).
We show in Sec.~\ref{sec:ablation} that removing the GRU
increases collision rates and jerk penalties, confirming temporal
aggregation is necessary for safe trajectories.

The output head produces a 6-dimensional vector that is reshaped into
an acceleration prediction $\mathbf{a}_{\mathrm{pred}}$ and an
auxiliary velocity estimate $\hat{\mathbf{v}}$, both in the
yaw-aligned body frame.
The velocity estimate provides a self-supervised training signal
(Sec.~\ref{sec:diff_physics}) and is discarded at deployment.

\subsection{Differentiable Physics Training}
\label{sec:diff_physics}

\begin{figure}[t]
  \centering
  \vspace{18pt}
  \includegraphics[width=\linewidth]{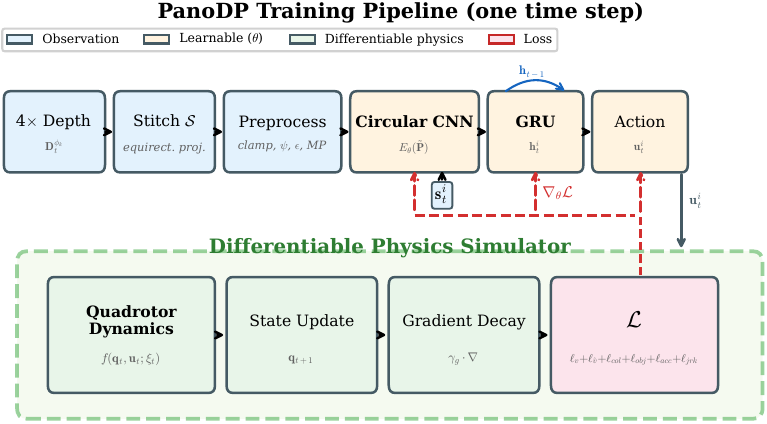}
  \vspace{-16pt}
  \caption{PanoDP training pipeline.
  At each time step the four onboard depth images are stitched into
  a $360^{\circ}$ panorama, encoded by the circular CNN, fused with
  the body-frame state vector in a GRU cell, and mapped to an
  acceleration command.
  Because the dynamics integrator~\eqref{eq:pos_update}--\eqref{eq:vel_update}
  is fully differentiable, the trajectory loss
  $\mathcal{L}$ can be back-propagated through the entire
  $T$-step unrolled graph (dashed arrows), yielding exact
  gradients $\partial\mathcal{L}/\partial\theta$ that update
  all learnable parameters in a single pass.}
  \label{fig:architecture}
    \vspace{-16pt}
\end{figure}

We train $\pi_\theta$ end-to-end by \emph{differentiating through
the physics simulator}.
Unlike reinforcement learning, which estimates policy gradients from
sampled returns, differentiable-physics training computes exact
analytic gradients $\partial\mathcal{L}/\partial\theta$ through
the full trajectory (see
Fig.~\ref{fig:architecture} for details).

\textbf{Differentiable rollout.}\;
The quadrotor is modeled as a second-order dynamical system.
At each control step the acceleration
$\mathbf{a}_{t}$ is determined by the policy output, aerodynamic
drag, and stochastic perturbations $\xi_{t}$; the state is then
integrated via
\begin{align}
\mathbf{p}_{t+1} &= \mathbf{p}_{t}+\mathbf{v}_{t}\,\Delta t
                    +\tfrac{1}{2}\,\mathbf{a}_{t}\,\Delta t^{2},
\label{eq:pos_update}\\
\mathbf{v}_{t+1} &= \mathbf{v}_{t}
                    +\tfrac{1}{2}(\mathbf{a}_{t}+\mathbf{a}_{t+1})\,\Delta t,
\label{eq:vel_update}
\end{align}
where $\Delta t$ is the control period.
The integrator and its Jacobians are implemented in
closed form as a differentiable operator, so the full $T$-step
rollout---from observation to panoramic encoding to action to next
state---forms a single computation graph.
Back-propagating the trajectory loss through this graph yields
analytic gradients
$\partial\mathcal{L}/\partial\theta$ that capture how each policy
parameter influences every loss term
($\ell_{v}$, $\ell_{\mathrm{col}}$, $\ell_{\mathrm{acc}}$, etc.)
across the horizon, in contrast to the high-variance
gradient estimates from reinforcement learning.

A gradient decay factor $\gamma_{g}\!\in\!(0,1)$ is applied at each
time step to attenuate gradients that flow across long horizons:
\begin{equation}
\frac{\partial\mathcal{L}}{\partial\mathbf{q}_{t}}
\;\leftarrow\;
\gamma_{g} \cdot
\frac{\partial\mathcal{L}}{\partial\mathbf{q}_{t}},
\label{eq:grad_decay}
\end{equation}
preventing gradient explosion while retaining the full  rollout.

\textbf{Training objective.}\;
Since every forward-simulation step is differentiable, the trajectory
loss can directly penalize undesirable behaviors---collisions,
goal-tracking errors, and jerky motion---across the full rollout
horizon.
Concretely, we minimize a weighted sum of per-step losses:
\begin{align}
\min_{\theta}\;&\;\mathbb{E}\!\left[\sum_{t=0}^{T-1}\ell(\mathbf{q}_{t},\mathbf{u}_{t})\right],
\label{eq:objective}\\
\ell\;=\;&\;\lambda_{v}\,\ell_{v}
       + \lambda_{\hat{v}}\,\ell_{\hat{v}}
       + \lambda_{\mathrm{col}}\,\ell_{\mathrm{col}}
\notag\\
       &+ \lambda_{\mathrm{obj}}\,\ell_{\mathrm{obj}}
       + \lambda_{\mathrm{acc}}\,\ell_{\mathrm{acc}}
       + \lambda_{\mathrm{jrk}}\,\ell_{\mathrm{jrk}}.
\label{eq:loss}
\end{align}
Each term is defined as follows:
\begin{itemize}
  \item $\ell_{v}$: Smooth-$L_1$ penalty on the velocity tracking error
        $\|\bar{\mathbf{v}}_{t}-\mathbf{v}^{*}_{t}\|$, where
        $\bar{\mathbf{v}}_t$ is a 30-step running-average velocity.
  \item $\ell_{\hat{v}}$: MSE loss between the auxiliary velocity
        prediction head $\hat{\mathbf{v}}_t$ and the true velocity,
        providing a self-supervised odometry signal.
  \item $\ell_{\mathrm{col}}$: proximity-weighted collision penalty
        $\mathrm{softplus}$ $(-\kappa\, d_{t})\cdot v_{\mathrm{app}}$,
        where $d_{t}$ is the margin-adjusted distance to the nearest
        neighbor and $v_{\mathrm{app}}$ is the approach speed.
  \item $\ell_{\mathrm{obj}}$: obstacle avoidance barrier
        $v_{\mathrm{app}}\cdot[\max(0,\,1\!-\!d_{t})]^{2}$,
        penalizing approach toward static obstacles.
  \item $\ell_{\mathrm{acc}}$: $L_2$ acceleration regularizer
        $\|\mathbf{a}_t\|^2$.
  \item $\ell_{\mathrm{jrk}}$: $L_2$ jerk regularizer
        $\|\dot{\mathbf{a}}_t\|^2$ (finite-difference).
\end{itemize}
The weighting coefficients $\{\lambda\}$ are reported in
Sec.~\ref{sec:experiments}.

\subsection{Random Rotation Augmentation}
\label{sec:random_rotation}
To prevent the policy from developing a dependence on a fixed global
heading, we apply a random yaw rotation $R_{\psi}$ (about the $z$-axis,
$\psi\sim\mathcal{U}[-\psi_{\max},+\psi_{\max}]$) to the
\emph{entire scene}---drone positions, goal positions, and all obstacle
coordinates---at each training episode reset.
Because every spatial quantity is rotated together, the physical
scenario is unchanged but its projection onto the panorama shifts
along the azimuth axis.
Across episodes this forces the circular encoder to treat all
bearing directions equally, complementing the circular padding
described in Sec.~\ref{sec:circular_conv}.
The value of $\psi_{\max}$ is reported in Sec.~\ref{sec:experiments}.


\section{Experiments}
\label{sec:experiments}

\noindent\textbf{Experimental setup.}
All learned policies are trained via differentiable physics
(Sec.~\ref{sec:diff_physics}) on a single NVIDIA RTX\,4090 GPU
(24\,GB VRAM) with batch size $B\!=\!1024$, unroll
horizon $T\!=\!165$ steps, learning rate $10^{-3}$ (AdamW, cosine
annealing), gradient decay factor $\gamma_{g}\!=\!0.4$, and the
following loss coefficients (Eq.~\eqref{eq:loss}):
$\lambda_{v}\!=\!1.0$,
$\lambda_{\hat{v}}\!=\!2.0$,
$\lambda_{\mathrm{col}}\!=\!5.0$,
$\lambda_{\mathrm{obj}}\!=\!2.0$,
$\lambda_{\mathrm{acc}}\!=\!0.01$,
$\lambda_{\mathrm{jrk}}\!=\!0.001$.
Random rotation augmentation ($\psi_{\max}\!=\!0.75$\,rad
$\approx\!43^{\circ}$, Sec.~\ref{sec:random_rotation})
is enabled for both
PanoDP and the forward-depth baseline DPD$^\dagger$.
Each training batch contains 4--8 agents;
the \emph{same} policy is deployed without retraining at
test time to swarms of up to 512+ agents.
PanoDP uses 514\,K parameters---nearly identical to
DPD$^\dagger$ (514\,K; the sole difference is the first-layer
kernel size), confirming that performance gains stem from the
panoramic representation rather than added model capacity.
The circular padding operation adds negligible computational
overhead over standard zero-padded convolutions, as it only
changes the fill mode of the same-size padding tensor.

\noindent\textbf{Metrics.}
We report three  metrics across all evaluations.
\begin{itemize}
  \item \textbf{Success Rate (SR\,$\uparrow$)}:
        the fraction of agents that complete the episode without any
        safety-margin violation,
        $\mathrm{SR}=N_{\mathrm{safe}}/N$.
        An agent is considered successful if its minimum clearance
        to all neighbors and obstacles remains positive throughout
        the entire episode.
    \item \textbf{Collision Rate (CR\,$\downarrow$)}:
        the number of new collision events (inter-agent and
        agent-obstacle margin violations) per second of simulation
        time, aggregated over all agents.
        A lower CR indicates safer swarm navigation.
    \item \textbf{Mean First Collision Time (MFCT\,$\uparrow$)}:
      the average time (in seconds) at which each colliding agent
      first violates the safety margin.
      If no collision occurs, MFCT equals the episode duration;
      higher is better.
\end{itemize}

\subsection{Training Ablations}
\label{sec:ablation}
We compare four architecture variants:
\begin{enumerate}
  \item \textbf{PanoDP (Ours)}: full system---panoramic depth,
        circular encoder, GRU memory.
  \item \textbf{w/o GRU}: Encoder without recurrent memory.
  \item \textbf{w/o Circular Conv}: standard zero-padded convolutions
        on the panoramic input.
  \item \textbf{Concat MLP}: panoramic views concatenated as a flat
        vector and passed through an MLP instead of a convolutional
        encoder.
\end{enumerate}

\begin{figure}[t]
  \centering
  \vspace{30pt}
  \includegraphics[width=\linewidth]{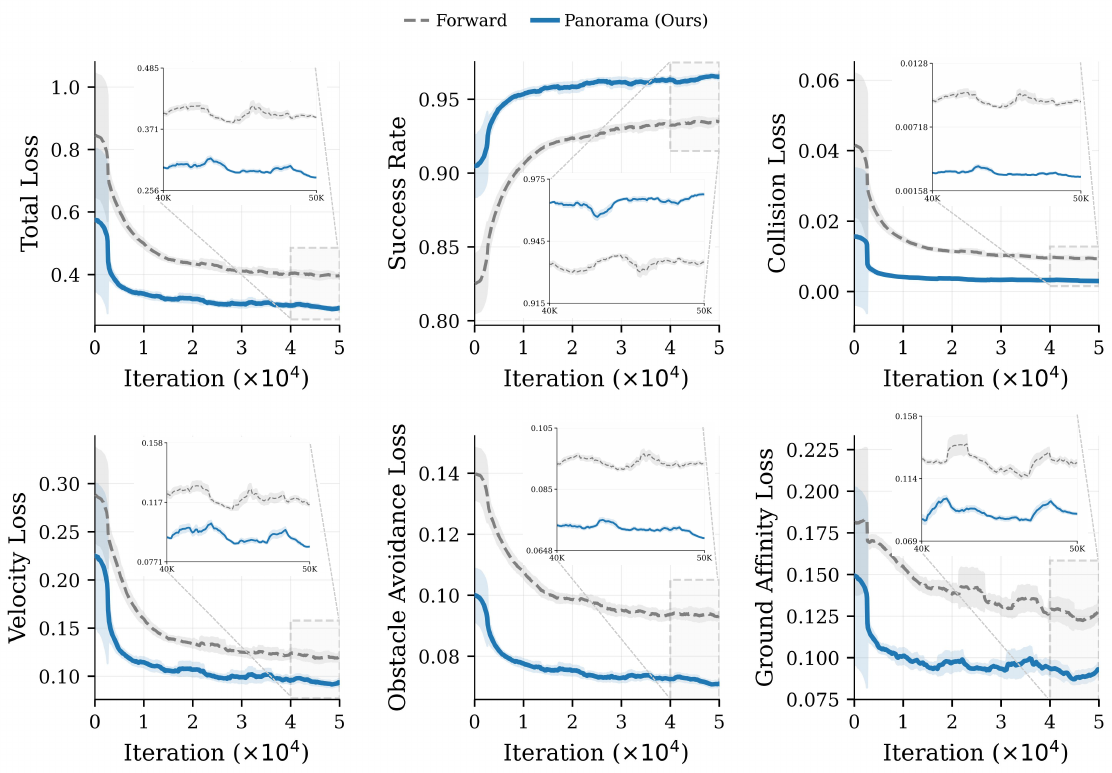}
    \vspace{-16pt}
  \caption{Panorama and forward-depth training comparison over 50\,K
  iterations (6 metrics). Insets zoom into the converged tail
  (40--50\,K steps).
  PanoDP (blue, solid) consistently outperforms the forward-only
  baseline DPD$^\dagger$ (grey, dashed) across all metrics.}
  \label{fig:training_pano_vs_forward}
    \vspace{-21pt}
\end{figure}

\begin{figure}[t]
  \centering
  \vspace{30pt}
  \includegraphics[width=\linewidth]{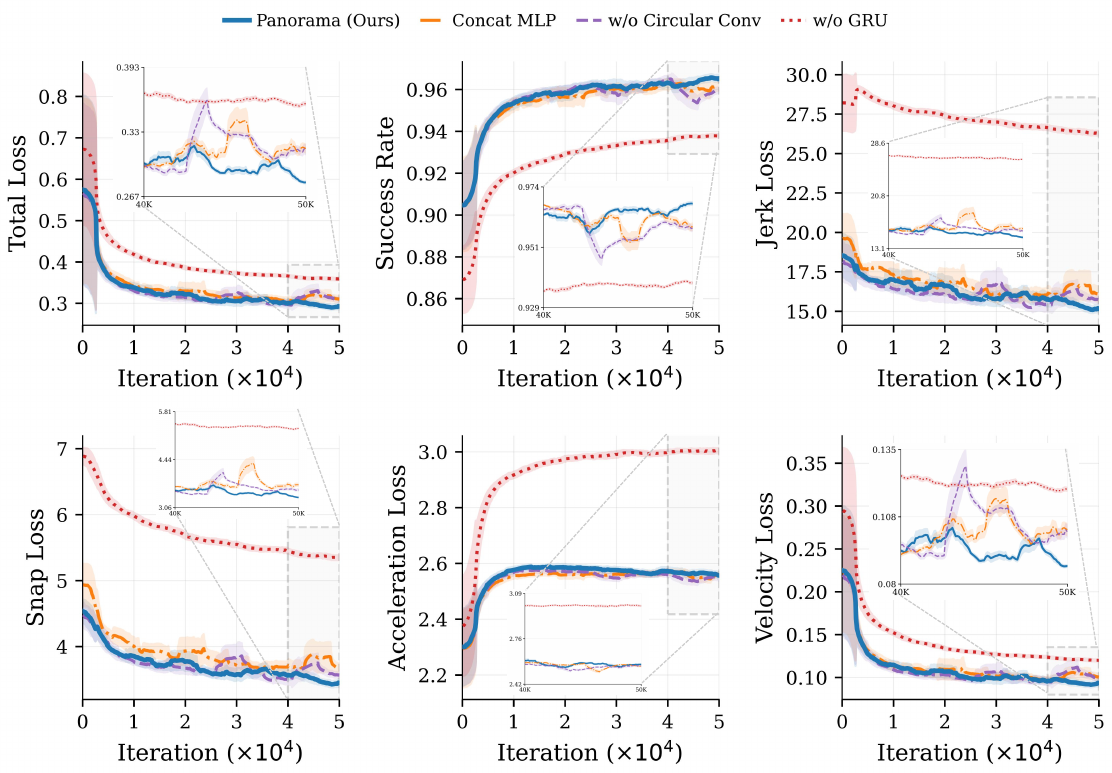}
  \vspace{-16pt}
\caption{Ablation training curves over 50\,K iterations (6 metrics), with tail zoom (40--50\,K).
Full PanoDP converges most stably with lowest variance.
Removing the GRU raises jerk and snap; removing circular convolutions or using a flat MLP yields higher plateaus.}
  \label{fig:training_ablation}
    \vspace{-10pt}
\end{figure}

\begin{figure}[t]
  \centering
  \includegraphics[width=\linewidth]{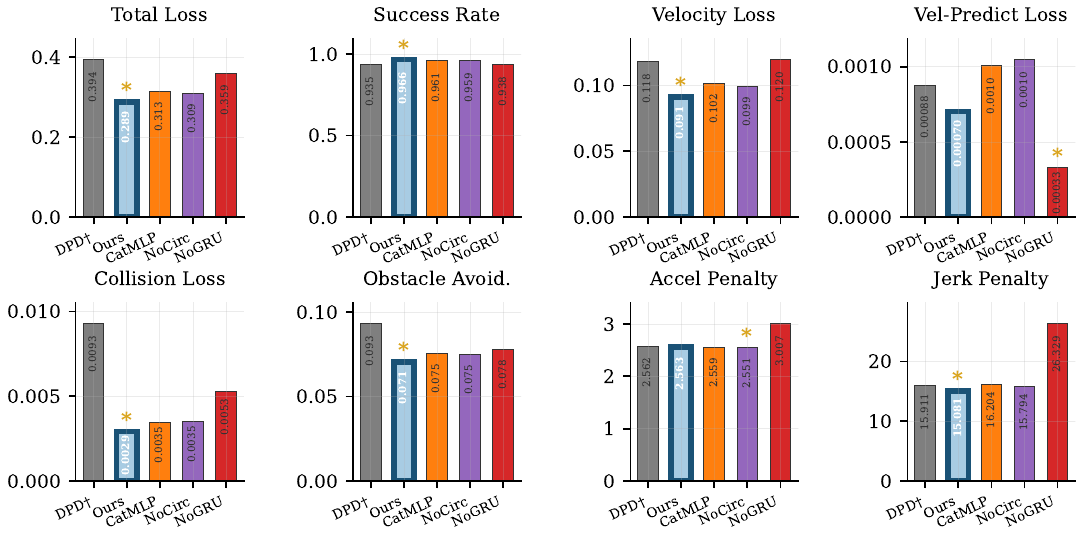}
  \vspace{-19pt}
  \caption{Ablation bar chart (tail-mean at 50\,K iterations) across
  eight training metrics.
  PanoDP (Ours, red border) achieves the best or near-best value
  on every metric.
  Stars ($\star$) mark the best method per subplot.}
  \label{fig:ablation_bars}
  \vspace{-19pt}
\end{figure}

Fig.~\ref{fig:training_pano_vs_forward} compares the full panoramic
policy against the forward-depth baseline;
Fig.~\ref{fig:training_ablation} presents the ablation training
curves; and Fig.~\ref{fig:ablation_bars} visualizes the tail-mean
metrics as bar charts.
The tail-mean values (averaged over the last~20 logged points up to
50\,K iterations) are given in Table~\ref{tab:ablation-50k}.
We highlight several key findings below.

\noindent\textbf{Benefit of panoramic coverage.}\;
PanoDP consistently outperforms DPD$^\dagger$ across all six
metrics (Fig.~\ref{fig:training_pano_vs_forward}).
Most notably, the collision loss drops by 68.5\% (0.003 vs.\ 0.009)
and the obstacle loss decreases by 24.3\% (0.071 vs.\ 0.093),
confirming that $360^{\circ}$ depth coverage enables the policy to
detect and avoid threats from all directions.
PanoDP also converges more smoothly and with lower variance;
it converges roughly twice as fast as
DPD$^\dagger$, indicating the panoramic signal provides a richer gradient landscape for
differentiable-physics optimizer.

\noindent\textbf{Role of GRU memory.}\;
Removing the GRU (\emph{w/o~GRU}) increases jerk by 74.6\% (26.33
vs.\ 15.08) and snap by 56.4\% (5.35 vs.\ 3.42; Table~\ref{tab:ablation-50k}), confirming that
the recurrent hidden state serves as a temporal smoother that
anticipates upcoming collisions from consecutive depth frames.
The success rate drops from 0.966 to 0.938 and the total loss rises
by 24.2\% (0.359 vs.\ 0.289), making \emph{w/o~GRU} the weakest variant.
Notably, \emph{w/o~GRU} achieves the lowest velocity-prediction
loss (Fig.~\ref{fig:ablation_bars}) yet produces the worst trajectory quality---without recurrent
memory the network trivially copies the current velocity,
forfeiting the multi-frame context needed for anticipatory control.

\noindent\textbf{Effect of circular convolution.}\;
Replacing circular convolutions with standard zero-padding
(\emph{w/o Circular Conv}) or collapsing the panorama into a flat MLP
input (\emph{Concat MLP}) raises the total loss by 7--8\% and
reduces the success rate to 0.959--0.961 (Table~\ref{tab:ablation-50k}).
Although these variants benefit from $360^{\circ}$ coverage,
lacking rotational equivariance introduces seam artifacts at the
panorama's $0^{\circ}/360^{\circ}$ seam, raising collision
and obstacle losses.
This validates the circular encoder 
(Sec.~\ref{sec:circular_conv}) as essential for fully exploiting
panoramic inputs.

\noindent\textbf{Overall.}\;
PanoDP achieves the best value on 9 of 11 tail-mean training metrics (Table~\ref{tab:ablation-50k}), including all safety-critical indicators (success rate, collision loss, obstacle loss) and all trajectory-quality indicators (jerk, snap). For the remaining two regularization terms (acceleration, gradient affinity), it is within 8\% of the best variant, indicating no meaningful trade-off. Each component---panoramic depth, circular convolution, and GRU memory---contributes measurably.

\begin{table*}[t]
\vspace{6pt}
\centering
\scriptsize
\caption{Training ablation at 50\,K iterations (tail-mean).
  Succ.: training rollout success rate $\in[0,1]$;
  AR: success-weighted mean speed (Succ.\,$\times$\,Speed);
  G.Aff.: mean squared positive altitude (ground affinity).}
\vspace{-2pt}
\label{tab:ablation-50k}
\input{tables/table_i_ablation_50k.tex}
\vspace{-19pt}
\end{table*}

\noindent\textbf{Selection of Baselines.}
To contextualize PanoDP's performance, we compare against seven
baselines spanning learning-based and classical paradigms.
All baselines are executed as decentralized controllers---each agent
computes its own control input using only locally available
information---in the same simulation environment and under identical
initial conditions.
Since these methods target diverse platforms and simulators,
we re-implement each in our GPU-batched simulation for controlled comparison:
\begin{itemize}
  \item \textbf{DPD$^\dagger$}~\cite{zhang2025learning}:
        our own forward-depth policy trained with the same infrastructure
        (same loss, same horizon, same random rotation), using a single
        ${\sim}79^{\circ}$ forward camera instead of a panorama---the most
        direct ablation of the panoramic component.
        Since the DPD framework~\cite{zhang2025learning} has already
        been shown to outperform PPO-based multi-agent RL alternatives,
        we adopt DPD$^\dagger$ as the primary learning-based reference
        and focus on the effect of observation modality.
  \item \textbf{D-CBF}~\cite{mestres2024distributed}:
        applies distributed CBF safety constraints with
        reciprocal effort sharing between neighbors.
  \item \textbf{RAMFT}~\cite{park2024resilient}:
        coordinates multi-agent motion via topological-neighbor-based
        flocking rules combined with repulsive obstacle avoidance.
  \item \textbf{CBF-AF}~\cite{gonccalves2024safe}:
        applies CBF constraints with adaptive decay rates
        for acceleration-level avoidance.
  \item \textbf{HIL-CBF}~\cite{fernandez2023distributed}:
        employs higher-order CBF constraints that account for
        acceleration-level dynamics.
  \item \textbf{DWA}~\cite{yasuda2023safe}:
        samples candidate acceleration directions and scores
        them against obstacle clearance and goal proximity.
        DWA and APF-CPP use internally fixed velocity parameters;
        their results in the speed panel
        (Tab.~\ref{tab:stress-test}\,c) are accordingly
        speed-invariant.
  \item \textbf{APF-CPP}~\cite{wang2024apf}:
        computes repulsive and attractive potential fields
        for reactive obstacle avoidance and goal reaching.
\end{itemize}
{\textit{Remark}.}\;
PanoDP operates from \emph{onboard observations}:
a panoramic depth image and a 10-D body-frame feature vector
(Sec.~\ref{sec:panoramic_depth}).
In contrast, all non-learning baselines receive \emph{privileged
state information}---exact positions and velocities of neighboring
agents within a sensing radius of $r_{s}\!=\!24$\,m (matching the
depth camera clamp range $d_{\max}$)---which would require
inter-agent communication or a motion-capture system in practice.
Despite this generous advantage, PanoDP---operating without
any inter-agent state exchange---matches or exceeds these
privileged methods in most configurations.
\subsection{Stress-Test Evaluation}
\label{sec:stress_test}
To measure robustness far beyond training conditions, we design a
comprehensive \emph{stress-test benchmark} using the challenging
circle-swap scenario, where all agents must simultaneously cross
through a congested central area to reach diametrically opposite
goal positions.
Each configuration is evaluated on 3 independently generated random
maps; all methods are tested on the identical map layouts under each
configuration to ensure a fair comparison.

We vary three axes independently:
\begin{enumerate}
  \item \emph{Swarm scale}: $N\!\in\!\{64,128,256,512\}$,
        at fixed obstacle density $\rho_{\mathrm{obs}}\!=\!1.0$ and
        randomized drone speed
        $v\!\sim\!\mathcal{U}[0.75,\,3.25]$\,m/s.
  \item \emph{Obstacle density}:
        $\rho_{\mathrm{obs}}\!\in\!\{0.5,1.0,1.5,2.0\}$,
        at $N\!=\!256$ and randomized speed
        $v\!\sim\!\mathcal{U}[0.75,\,3.25]$\,m/s.
  \item \emph{Drone speed}: $v\!\in\!\{0.75,1.5,2.25,3.0\}$~m/s,
        at $N\!=\!256$, $\rho_{\mathrm{obs}}\!=\!1.0$, and static
        obstacles.
\end{enumerate}

Table~\ref{tab:stress-test} presents the full results.

\begin{table*}[t]
\vspace{6pt}
\centering
\scriptsize
\caption{Stress-test benchmark (circle-swap): three-panel comparison
across 12\,configurations (3 maps each).
\textbf{Bold} = best; \underline{underline} = second.
Non-learning baselines receive privileged neighbor state
($r_s\!=\!24$\,m); PanoDP uses only onboard depth.}
\vspace{-2pt}
\label{tab:stress-test}
\input{tables/table_threepanel_round2_12cfg.tex}

\vspace{-18pt}
\end{table*}

\noindent \textbf{Swarm scale}
(Tab.~\ref{tab:stress-test}\,a).\;
As $N$ increases from 64 to 512, central congestion intensifies.
PanoDP sustains a high success rate (87.2\% at $N\!=\!512$) while
maintaining the lowest collision rate among learned methods.
The forward baseline DPD$^\dagger$, lacking lateral perception,
only reaches 62.2\% SR with CR~3.05 at $N\!=\!512$---25 percentage
points below PanoDP, confirming that blind-spot collisions dominate
in dense swarms.
Even the strongest privileged baseline D-CBF (84.4\% SR) cannot match
PanoDP at this extreme.
PanoDP also achieves the highest MFCT across all scales (e.g.\
48.6\,s at $N\!=\!512$ vs.\ 40.4\,s for D-CBF), indicating that its
collisions, when they occur, happen later in the episode.

\noindent \textbf{Obstacle density}
(Tab.~\ref{tab:stress-test}\,b).\;
PanoDP retains stable performance across densities
(SR 80.5\%$\to$79.6\% as $\rho_{\mathrm{obs}}$ doubles from 0.5
to 2.0), while consistently leading in MFCT (36.6--40.1\,s).
Classical planners exhibit sharper degradation; e.g., APF-CPP drops
from 32.8\% to 22.5\%.

\noindent \textbf{Drone speed}
(Tab.~\ref{tab:stress-test}\,c).\;
At the highest speed ($v\!=\!3.0$\,m/s), reaction time shrinks and
the cost of a missed obstacle is severe.
PanoDP achieves 76.6\% SR vs.\ 44.5\% for DPD$^\dagger$,
showing that $360^{\circ}$ awareness is especially valuable when the
agent cannot afford to ``look back."
Compared with the best privileged baseline at this speed (D-CBF,
73.3\%), PanoDP still leads by 3.3 percentage points despite
operating without any inter-agent state exchange.
Across all three speed levels, PanoDP maintains the highest MFCT,
confirming that $360^{\circ}$ awareness delays collisions more
effectively.

\subsection{Robustness to Single-Camera Failure}
\label{sec:occlusion}

\textbf{Motivation.}\;
On a real platform any depth camera may suffer sudden obstruction or
hardware failure.
To quantify graceful degradation, we replace each camera's
$64{\times}64$ depth image with a max-range reading ($d{=}100$\,m)
before panoramic stitching, simulating a fully failed sensor.
Each condition is evaluated over 10~seeds $\times$ 1024~drones
($T{=}165$ steps).
Crucially, the obstacle layout is \emph{geometrically symmetric}:
obstacles are uniformly sampled across the arena and the eight
start--goal pairs are mirror-symmetric about the longitudinal axis,
so any left--right performance asymmetry cannot stem from
environmental bias.

\begin{figure}[t]
\centering
\includegraphics[width=\linewidth]{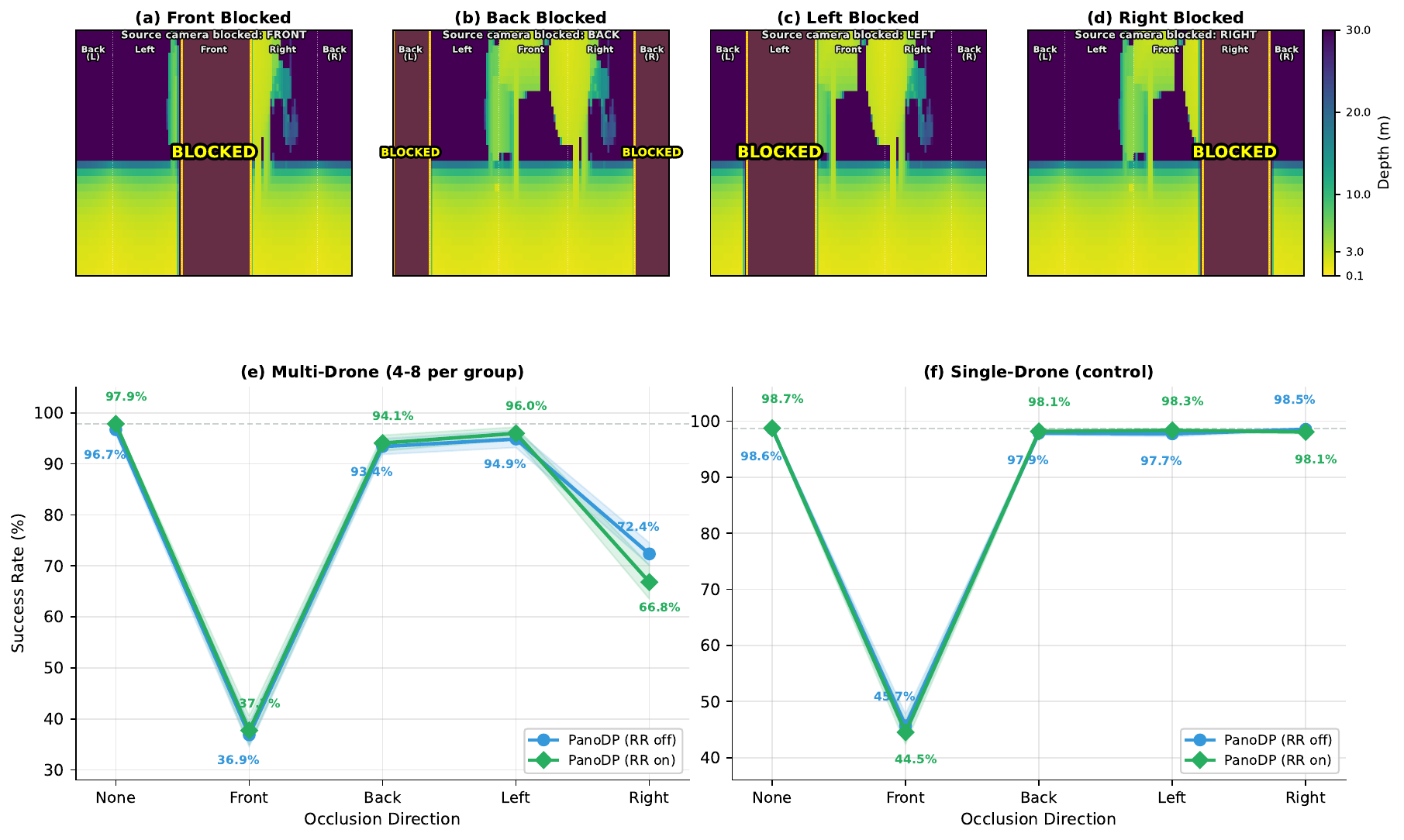}
\vspace{-16pt}
\caption{Single-camera occlusion robustness.
\textbf{(a)--(d)} 360\textdegree{} panoramic depth with one camera blocked (yellow: max-range fill).
\textbf{(e)} In multi-drone tests, blocking the right camera drops SR by 24 pp, while blocking the left has negligible effect, revealing a directional convention.
\textbf{(f)} In single-drone control, any non-front camera is expendable ($<$1 pp drop), showing the asymmetry stems from coordination.
Shaded bands: $\pm$1 std over 10 seeds.}
\label{fig:occ_combined}
\vspace{-10pt}
\end{figure}

\textbf{Discovery.}\;
In multi-drone mode (Fig.~\ref{fig:occ_combined}e), front-camera
failure causes a catastrophic $-59.8$~pp drop ($96.7\%{\to}36.9\%$),
as expected for the primary look-ahead sensor.
Surprisingly, \emph{right}-camera occlusion incurs $-24.3$~pp while
\emph{left}-camera loss causes only $-1.8$~pp---a stark directional
asymmetry in a geometrically symmetric environment.

\textbf{Verification.}\;
A single-drone control (Fig.~\ref{fig:occ_combined}f) removes
inter-drone collisions entirely.
Blocking any non-front camera now causes $<$1~pp degradation, and
left vs.\ right cameras become equally expendable
($97.7\%$ vs.\ $98.5\%$).
The right-camera asymmetry therefore arises exclusively from
multi-agent interactions: the policy has learned an emergent
\emph{right-hand traffic convention}---when drones meet head-on,
they consistently veer right, making the right camera critical for
detecting  along the preferred avoidance direction.
This emergent behavior is visually confirmed in the circle-swap
demonstration (Fig.~\ref{fig:demo_circle_swap}), where the swarm
exhibits a coherent counterclockwise rotation pattern as agents
cross the congested center via right-side avoidance.
Since random rotation augmentation is symmetric
($\psi\!\sim\!\mathcal{U}[-\psi_{\max},\psi_{\max}]$) and
DPD$^\dagger$ trained with the same augmentation shows no such
directional preference, the convention is attributable to panoramic
coverage rather than training-time rotational bias.

This self-organized convention,enabled by $360^{\circ}$
panoramic awareness acting as an implicit spatial communication
channel, explains the strong performance of PanoDP in the circle-swap
stress test (Sec.~\ref{sec:stress_test}), where
coordinated crossing through a congested center would be impossible
without such emergent traffic rules.

\subsection{Qualitative Demonstrations}
\label{sec:demo}

\begin{figure}[t]
\vspace{8pt}
  \centering
  \includegraphics[width=\linewidth]{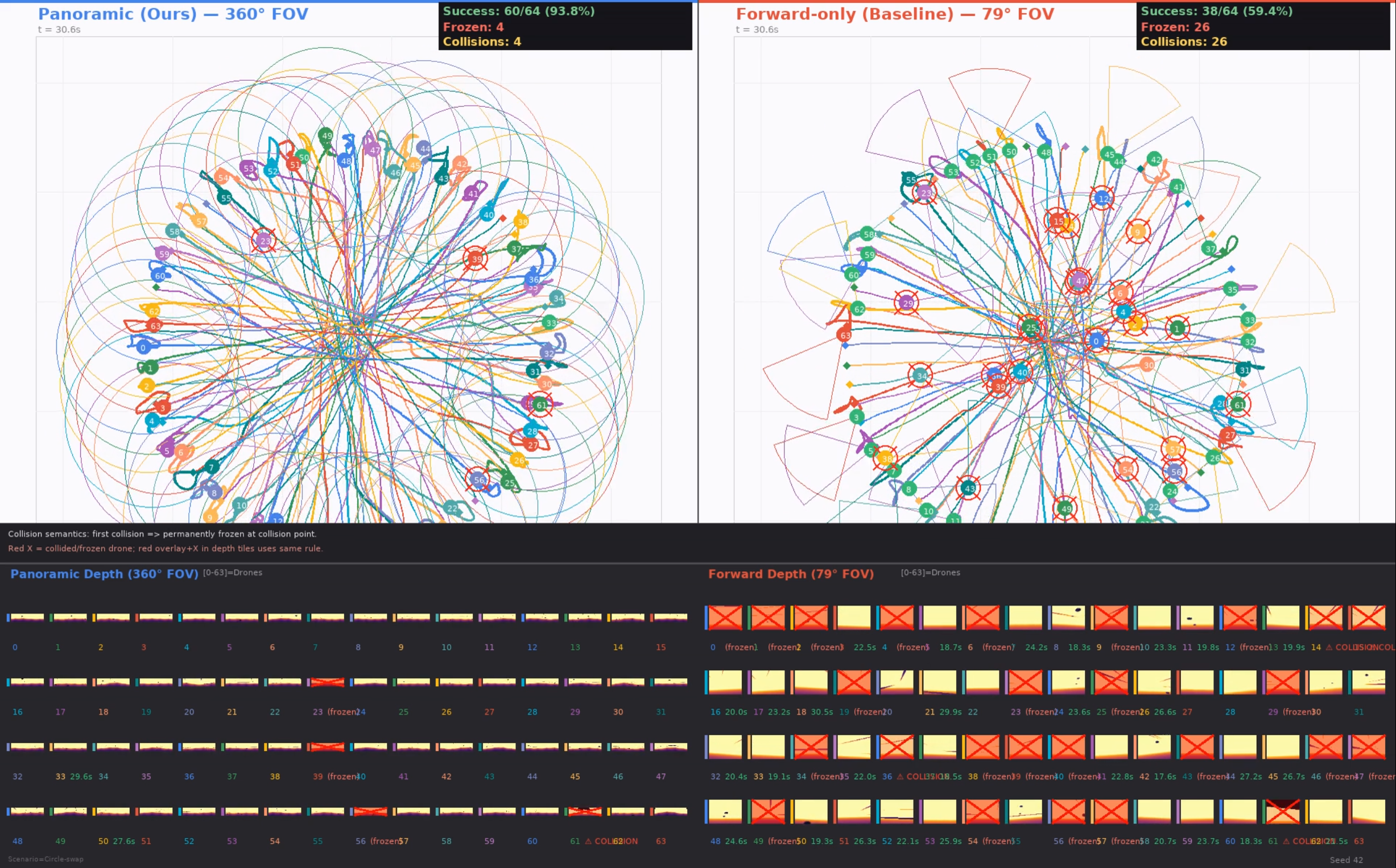}
  \vspace{-15pt}
  \caption{Circle-swap with $N\!=\!64$.
  Left: Panoramic (93.8\% success, 4 collisions);
  Right: Forward-only (59.4\%, 26 collisions).
  The swarm self-organizes a counterclockwise rotation through
  the congested center.
  Bottom: per-drone depth tiles; red \textsf{X} = frozen after collision.}
  \label{fig:demo_circle_swap}
  \vspace{-18pt}
\end{figure}

\noindent\textbf{Circle-swap}
(Fig.~\ref{fig:demo_circle_swap}).\;
With $N\!=\!64$ agents crossing through a single congested center,
Panoramic achieves 93.8\% success (4 collisions) versus 59.4\%
(26 collisions) for Forward-only.
The overhead view reveals an emergent \emph{counterclockwise
rotation}---each agent veers right that corroborates the occlusion analysis
(Sec.~\ref{sec:occlusion}).

\begin{figure}[t]
\vspace{15pt}
  \centering
  \includegraphics[width=\linewidth]{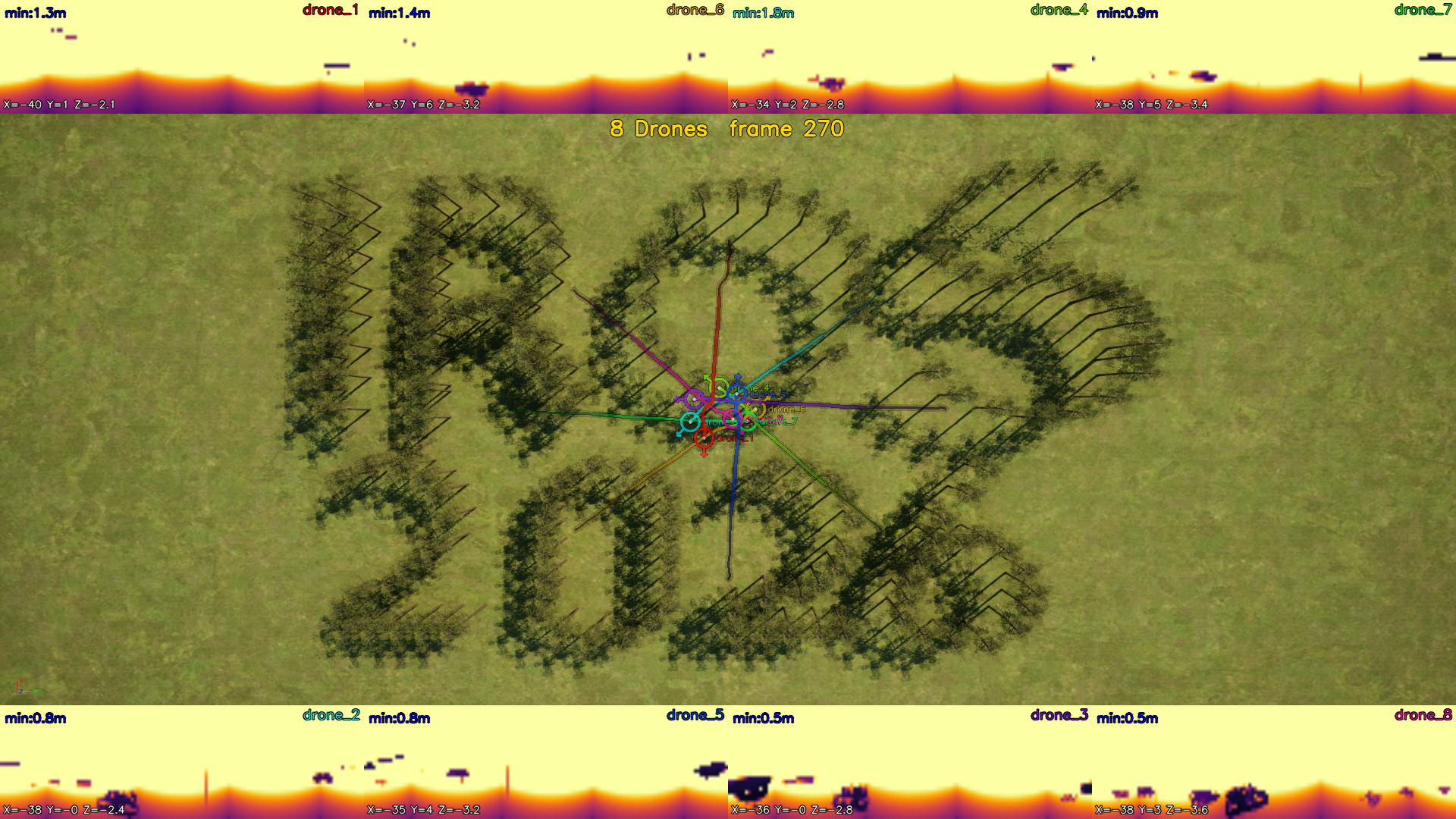}
  \vspace{-15pt}
\caption{Sim-to-sim transfer in AirSim: a multi-drone swarm navigates a dense bamboo forest with photorealistic depth and stochastic wind. A policy trained in a lightweight differentiable simulator with spherical obstacles transfers to AirSim without retraining, showing robustness to visual domain shift and aerodynamic perturbations.}
  \label{fig:demo_bamboo}
  \vspace{-18pt}
\end{figure}

\noindent\textbf{Bamboo-forest traversal}
(Fig.~\ref{fig:demo_bamboo}).\;
Deploying the same policy in an AirSim~\cite{shah2017airsim}
bamboo forest, where depth images contain thin-structure geometry
never seen during training and stochastic wind introduces forces
absent from training, the swarm successfully navigates the forest.
This suggests that inverse-depth preprocessing and GRU temporal
memory (Sec.~\ref{sec:gru_policy}) jointly provide sufficient
robustness to bridge the simulator-to-deployment gap.

\section{Conclusion}
\label{sec:conclusion}

We presented PanoDP, a decentralized collision-avoidance policy for large quadrotor swarms that runs purely on onboard $360^{\circ}$ panoramic depth. It combines an equirectangular panoramic representation, a circular convolutional encoder for azimuthal wrap-around, and a GRU module for temporal memory and implicit velocity estimation. Trained end-to-end with differentiable physics using only 4--8 agents per batch on a single GPU, the same policy generalizes to 1 to 512+ agents at test time without retraining or communication, with per-agent $O(1)$ cost.
Across out-of-distribution circle-swap stress tests, PanoDP matches or exceeds privileged planners and reaches 87.2\% collision-free rate at $N\!=\!512$. Occluding a single camera reveals an emergent right-hand traffic convention, also visible in the circle-swap visualization. Sim-to-sim transfer to a photorealistic AirSim bamboo forest further shows robustness to visual domain shift.

\textbf{Limitations and future work.}\;
The current evaluation considers spherical static obstacles and speeds up to $3.25$\,m/s. Extending to dynamic obstacles and higher speeds will further stress-test the panoramic encoder. Scaling beyond 512 agents and higher panoramic resolution likely requires multi-GPU training, while a fused CUDA stitching kernel can reduce per-frame overhead for onboard inference. Future work targets transfer to physical multi-drone platforms under sensor noise, latency, and regulatory constraints.

\bibliographystyle{IEEEtran}
\bibliography{ref}

\end{document}

%% file: tables/table_i_ablation_50k.tex
\begin{tabular}{lccccccccccc}
\hline\hline
Method & Loss$\downarrow$ & Succ.$\uparrow$ & AR$\uparrow$ & Col.$\downarrow$ & $\mathcal{L}_v\downarrow$ & $\mathcal{L}_{obj}\downarrow$ & Jerk$\downarrow$ & Snap$\downarrow$ & Acc.$\downarrow$ & G.Aff.$\downarrow$ & Speed$\uparrow$ \\
\hline
DPD$^\dagger$ (Forward) & 0.394 & 0.935 & 1.479 & 0.009 & 0.118 & 0.093 & 15.9110 & 3.6412 & 2.5618 & 0.1256 & 1.579 \\
PanoDP (Ours) & \textbf{0.289} & \textbf{0.966} & \textbf{1.558} & \textbf{0.003} & \textbf{0.091} & \textbf{0.071} & \textbf{15.0810} & \textbf{3.4216} & 2.5628 & 0.0912 & \textbf{1.617} \\
\hline
w/o GRU & 0.359 & 0.938 & 1.491 & 0.005 & 0.120 & 0.078 & 26.3286 & 5.3502 & 3.0066 & 0.1145 & 1.597 \\
w/o Circular Conv & 0.309 & 0.959 & 1.535 & 0.004 & 0.099 & 0.075 & 15.7935 & 3.5793 & \textbf{2.5514} & \textbf{0.0851} & 1.605 \\
Concat MLP & 0.313 & 0.961 & 1.533 & 0.003 & 0.102 & 0.075 & 16.2036 & 3.6878 & 2.5590 & 0.0892 & 1.601 \\
\hline\hline
\end{tabular}

%% file: tables/table_threepanel_round2_12cfg.tex

\begin{tabular}{l|ccc|ccc|ccc|ccc}
\hline\hline
\multicolumn{13}{c}{\textbf{(a) Scale Robustness ($\rho_{\mathrm{obs}}\!=\!1.0$, $v_{\mathrm{drone}}\!=\!\mathrm{rand}$)}} \\
\hline
Method & \multicolumn{3}{c}{$N\!=\!64$} & \multicolumn{3}{c}{$N\!=\!128$} & \multicolumn{3}{c}{$N\!=\!256$} & \multicolumn{3}{c}{$N\!=\!512$} \\
 & SR($\%$)$\uparrow$ & CR$\downarrow$ & MFCT(s)$\uparrow$ & SR($\%$)$\uparrow$ & CR$\downarrow$ & MFCT(s)$\uparrow$ & SR($\%$)$\uparrow$ & CR$\downarrow$ & MFCT(s)$\uparrow$ & SR($\%$)$\uparrow$ & CR$\downarrow$ & MFCT(s)$\uparrow$ \\
\hline
PanoDP (Ours) & \underline{84.4} & \underline{0.27} & \textbf{16.8} & \textbf{80.7} & \textbf{0.38} & \textbf{32.5} & \textbf{80.6} & \underline{0.80} & \textbf{37.5} & \textbf{87.2} & \textbf{0.88} & \textbf{48.6} \\
D-CBF~\cite{mestres2024distributed} & 82.8 & 0.30 & 9.1 & \underline{78.1} & \underline{0.43} & 19.9 & \underline{78.6} & \textbf{0.76} & 25.0 & \underline{84.4} & \underline{1.03} & 40.4 \\
RAMFT~\cite{park2024resilient} & \textbf{86.5} & \textbf{0.24} & \underline{10.3} & 74.7 & 0.46 & \underline{27.4} & 74.2 & 0.83 & \underline{31.3} & 79.4 & 1.32 & 33.6 \\
CBF-AF~\cite{gonccalves2024safe} & 67.7 & 0.60 & 7.7 & 66.9 & 0.61 & 17.0 & 67.7 & 1.04 & 24.9 & 74.0 & 1.81 & 40.8 \\
HIL-CBF~\cite{fernandez2023distributed} & 67.2 & 0.52 & 6.3 & 51.8 & 1.30 & 16.8 & 54.9 & 2.14 & 25.1 & 57.0 & 5.16 & \underline{41.5} \\
DWA~\cite{yasuda2023safe} & 62.0 & 0.65 & 3.9 & 60.2 & 0.73 & 5.9 & 60.3 & 1.28 & 7.4 & 62.6 & 2.50 & 9.4 \\
APF-CPP~\cite{wang2024apf} & 39.6 & 2.14 & 2.5 & 22.7 & 3.72 & 3.0 & 25.3 & 2.42 & 4.0 & 23.1 & 6.49 & 6.0 \\
DPD$^\dagger$~\cite{zhang2025learning} & 50.0 & 0.98 & 4.1 & 51.8 & 1.01 & 7.7 & 55.1 & 1.68 & 15.5 & 62.2 & 3.05 & 20.8 \\
\hline
\multicolumn{13}{c}{\textbf{(b) Obstacle Density ($N\!=\!256$, $v_{\mathrm{drone}}\!=\!\mathrm{rand}$)}} \\
\hline
Method & \multicolumn{3}{c}{$\rho_{\mathrm{obs}}\!=\!0.5$} & \multicolumn{3}{c}{$\rho_{\mathrm{obs}}\!=\!1.0$} & \multicolumn{3}{c}{$\rho_{\mathrm{obs}}\!=\!1.5$} & \multicolumn{3}{c}{$\rho_{\mathrm{obs}}\!=\!2.0$} \\
 & SR($\%$)$\uparrow$ & CR$\downarrow$ & MFCT(s)$\uparrow$ & SR($\%$)$\uparrow$ & CR$\downarrow$ & MFCT(s)$\uparrow$ & SR($\%$)$\uparrow$ & CR$\downarrow$ & MFCT(s)$\uparrow$ & SR($\%$)$\uparrow$ & CR$\downarrow$ & MFCT(s)$\uparrow$ \\
\hline
PanoDP (Ours) & \underline{80.5} & 0.80 & \textbf{36.6} & \textbf{80.6} & \underline{0.80} & \textbf{37.5} & \textbf{83.6} & \textbf{0.57} & \textbf{36.7} & \textbf{79.6} & 0.89 & \textbf{40.1} \\
D-CBF~\cite{mestres2024distributed} & \textbf{82.0} & \textbf{0.72} & 25.3 & \underline{78.6} & \textbf{0.76} & 25.0 & \underline{75.8} & 0.87 & 27.0 & \underline{75.8} & \underline{0.88} & 26.4 \\
RAMFT~\cite{park2024resilient} & 77.3 & \underline{0.73} & \underline{31.5} & 74.2 & 0.83 & \underline{31.3} & 73.8 & \underline{0.87} & \underline{30.4} & 72.9 & \textbf{0.87} & \underline{30.6} \\
CBF-AF~\cite{gonccalves2024safe} & 75.3 & 1.04 & 23.4 & 67.7 & 1.04 & 24.9 & 62.4 & 1.30 & 23.7 & 63.9 & 1.18 & 23.4 \\
HIL-CBF~\cite{fernandez2023distributed} & 62.2 & 1.79 & 26.1 & 54.9 & 2.14 & 25.1 & 50.8 & 2.35 & 24.6 & 50.4 & 2.66 & 24.4 \\
DWA~\cite{yasuda2023safe} & 63.3 & 1.18 & 7.7 & 60.3 & 1.28 & 7.4 & 59.4 & 1.30 & 6.9 & 59.4 & 1.30 & 6.9 \\
APF-CPP~\cite{wang2024apf} & 32.8 & 2.20 & 4.4 & 25.3 & 2.42 & 4.0 & 20.8 & 4.99 & 3.8 & 22.5 & 3.75 & 3.6 \\
DPD$^\dagger$~\cite{zhang2025learning} & 58.1 & 1.68 & 13.4 & 55.1 & 1.68 & 15.5 & 53.6 & 1.77 & 15.0 & 58.3 & 1.52 & 13.4 \\
\hline
\multicolumn{13}{c}{\textbf{(c) Drone Speed ($N\!=\!256$, $\rho_{\mathrm{obs}}\!=\!1.0$, static obstacles)}} \\
\hline
Method & \multicolumn{3}{c}{$v\!=\!0.75$} & \multicolumn{3}{c}{$v\!=\!1.5$} & \multicolumn{3}{c}{$v\!=\!2.25$} & \multicolumn{3}{c}{$v\!=\!3.0$} \\
 & SR($\%$)$\uparrow$ & CR$\downarrow$ & MFCT(s)$\uparrow$ & SR($\%$)$\uparrow$ & CR$\downarrow$ & MFCT(s)$\uparrow$ & SR($\%$)$\uparrow$ & CR$\downarrow$ & MFCT(s)$\uparrow$ & SR($\%$)$\uparrow$ & CR$\downarrow$ & MFCT(s)$\uparrow$ \\
\hline
PanoDP (Ours) & \textbf{96.7} & \textbf{0.23} & 15.7 & \textbf{86.1} & \underline{0.54} & \textbf{40.8} & \underline{78.5} & \underline{0.91} & \textbf{36.4} & \textbf{76.6} & \textbf{1.23} & \textbf{27.3} \\
D-CBF~\cite{mestres2024distributed} & \underline{91.3} & \underline{0.45} & \underline{42.5} & 79.4 & 0.87 & 31.7 & 75.8 & 1.00 & 22.9 & \underline{73.3} & \underline{1.32} & \underline{17.1} \\
RAMFT~\cite{park2024resilient} & 83.6 & 0.52 & 36.8 & \underline{83.6} & \textbf{0.52} & \underline{40.4} & \textbf{80.7} & \textbf{0.68} & \underline{28.5} & 69.1 & 1.46 & 16.7 \\
CBF-AF~\cite{gonccalves2024safe} & 84.6 & 0.49 & \textbf{47.6} & 66.0 & 1.18 & 30.8 & 57.0 & 1.65 & 21.7 & 51.8 & 2.25 & 15.5 \\
HIL-CBF~\cite{fernandez2023distributed} & 61.1 & 4.07 & 41.5 & 52.7 & 2.31 & 29.7 & 51.3 & 2.09 & 22.0 & 49.7 & 2.56 & 15.4 \\
DWA~\cite{yasuda2023safe} & 60.3 & 1.28 & 7.4 & 60.3 & 1.28 & 7.4 & 60.3 & 1.40 & 7.4 & 60.4 & 1.85 & 7.2 \\
APF-CPP~\cite{wang2024apf} & 25.3 & 2.42 & 4.0 & 25.3 & 2.42 & 4.0 & 25.5 & 2.65 & 3.8 & 25.5 & 3.50 & 3.8 \\
DPD$^\dagger$~\cite{zhang2025learning} & 69.9 & 1.42 & 18.0 & 56.2 & 1.82 & 20.3 & 52.6 & 2.04 & 13.7 & 44.5 & 2.80 & 10.6 \\
\hline\hline
\end{tabular}